\documentclass[11pt, a4paper]{article}
\usepackage[round]{natbib}

\usepackage[normalem]{ulem}
\usepackage{mathtools}
\usepackage{setspace} 
\usepackage{dsfont}
\usepackage{amsfonts}
\usepackage{amsmath}
\usepackage{subcaption}
\usepackage{paralist}
\usepackage{times}
\usepackage{latexsym}
\usepackage{graphicx}
\usepackage[T1]{fontenc}
\usepackage{tikz}
\usepackage{url}
\usepackage{pgfplotstable}
\usepackage{titlesec}
\usepackage{color}
\usepackage{lipsum,adjustbox}
\usepackage[font={small}]{caption}
\usetikzlibrary{positioning}
\usepackage{bbm}
\usepackage{booktabs}
\usepackage{numprint}
\usepackage{siunitx}
\usepackage{xr}
\usepackage{placeins}

\makeatletter
\newcommand{\@BIBLABEL}{\@emptybiblabel}
\newcommand{\@emptybiblabel}[1]{}
\usepackage[hidelinks]{hyperref}

\usepackage{acl2018}
\graphicspath{{./plots/}}
\newcommand{\com}[1]{}

\newcommand{\maege}{{\sc maege}}

\newenvironment{myequation*}{
	\vspace{-.7em}
	\begin{equation*}
}{
\end{equation*}
\vspace{-.7em}
}

\newcommand{\my}[1]{}

\title{Automatic Metric Validation for Grammatical Error Correction}
\aclfinalcopy 

\begin{document}

\author{
  Leshem Choshen\textsuperscript{1} and Omri Abend\textsuperscript{1,2} \\
 \textsuperscript{1}School of Computer Science and Engineering,
  \textsuperscript{2} Department of Cognitive Sciences \\
  The Hebrew University of Jerusalem \\
  \texttt{leshem.choshen@mail.huji.ac.il, oabend@cs.huji.ac.il}\\
}

\maketitle
\begin{abstract}
  Metric validation in Grammatical Error Correction (GEC) is currently done by observing
  the correlation between human and metric-induced rankings. 
	However, such correlation studies are costly, 
  methodologically troublesome, and suffer from low inter-rater agreement.
  We propose \maege, an automatic methodology for GEC metric validation,
  that overcomes many of the difficulties with existing practices.
  Experiments with \maege\ shed a new light on metric quality, 
  showing for example that the standard $M^2$ metric fares poorly on corpus-level ranking.
	Moreover, we use \maege\ to perform a detailed analysis of metric
	behavior, showing that correcting some types of errors is consistently
  penalized by existing metrics.

\end{abstract}

\section{Introduction}

Much recent effort has been devoted to automatic evaluation, both within GEC \cite[see \S \ref{sec:measures}]{napoles2015ground, felice2015towards, ng2014conll, dahlmeier2012better}, and more generally in text-to-text generation tasks.
Within Machine Translation (MT), an annual shared task is devoted to automatic metric development,
accompanied by an extensive analysis of metric behavior \cite{bojar2017findings}.
Metric validation is also raising interest in GEC, with several recent works on the subject
\citep{grundkiewicz2015human,napoles2015ground,napoles-sakaguchi-tetreault:2016:EMNLP2016,sakaguchi2016reassessing},
all using correlation with human rankings (henceforth, {\it CHR}) as their methodology.

Human rankings are often considered as ground truth in text-to-text generation,
but using them reliably can be challenging.
Other than the costs of compiling a sizable validation set,
human rankings are known to yield poor inter-rater agreement in MT
\cite{bojar2011grain, lopez2012putting, graham2012measurement}, 
and to introduce a number of methodological problems that are difficult to overcome, 
notably the treatment of ties in the rankings and uncomparable sentences (see \S\ref{sec:Human judgments}).
These difficulties have motivated several proposals to alter the MT metric
validation protocol \cite{koehn2012simulating,dras2015evaluating},
leading to a recent abandoning of evaluation by human rankings due to its unreliability
\cite{graham2015accurate, bojar-EtAl:2016:WMT2}. 
These conclusions have not yet been implemented in GEC, despite their relevance.
In \S\ref{sec:Human judgments} we show that human rankings in GEC also suffer 
from low inter-rater agreement, motivating the development of alternative methodologies.

The main contribution of this paper is an automatic methodology for metric validation in GEC
called \maege\ (Methodology for Automatic Evaluation of GEC Evaluation), which addresses these difficulties.
\maege\ requires no human rankings, and instead uses a corpus with gold standard GEC annotation to generate 
lattices of corrections with similar meanings but varying degrees of grammaticality.
For each such lattice, \maege\ generates a partial order of correction quality,
a quality score for each correction, 
and the number and types of edits required to fully correct each. 
It then computes the correlation of the induced partial order with the metric-induced rankings.

\maege\  addresses many of the problems with existing methodology:
\begin{itemize}
	\vspace{-0.3cm}
 \item
  Human rankings yield low inter-rater and intra-rater agreement (\S\ref{sec:Human judgments}).
  Indeed, \citet{choshen2018conservatism} show that while annotators often generate different corrections
  given a sentence, they generally agree on whether a correction is valid or not. Unlike CHR, \maege\ bases its
  scores on human corrections, rather than on rankings.
\item
  CHR uses system outputs to obtain human rankings, which may be misleading, as systems may share similar biases, 
  thus neglecting to evaluate some types of valid corrections (\S\ref{sec:types}). 
  \maege\ addresses this issue by systematically traversing an inclusive space of corrections.
\item
  The difficulty in handling ties is addressed by only evaluating correction pairs
  where one contains a sub-set of the errors of the other, and is therefore clearly better.
\item
  \maege\ uses established statistical tests for determining the significance of its results, 
  thereby avoiding ad-hoc methodologies used in CHR to tackle potential biases in human rankings (\S\ref{sec:corpus}, \S\ref{sec:sentence}).
\end{itemize}


In experiments on the standard NUCLE test set \cite{dahlmeier2013building}, we find that \maege\ often disagrees with CHR
as to the quality of existing metrics. For example, we find that the standard GEC metric, $M^2$, is a poor predictor
of corpus-level ranking, but a good predictor of sentence-level pair-wise rankings.
The best predictor of corpus-level quality by \maege\ is the 
reference-less LT metric \cite{milkowski2010developing,napoles-sakaguchi-tetreault:2016:EMNLP2016},
while of the reference-based metrics, GLEU \cite{napoles2015ground} fares best.

In addition to measuring metric reliability, \maege\ 
can also be used to analyze the sensitivities of the metrics 
to corrections of different types, which to our knowledge is a novel contribution of this work.
Specifically, we find that not only are valid edits of some
error types better rewarded than others, but that correcting certain error types is
consistently penalized by existing metrics (Section \ref{sec:types}). 
The importance of interpretability and detail in evaluation practices
(as opposed to just providing bottom-line figures), 
has also been  stressed in MT evaluation \cite[e.g.,][]{birch2016hume}.

\section{Examined Metrics}\label{sec:measures}

We turn to presenting the metrics we experiment with.
The standard practice in GEC evaluation is to define differences between the source and a correction (or a reference) as a set of edits \cite{dale2012hoo}.
An edit is a contiguous span of tokens to be edited, a substitute string, and the corrected error type.
For example: ``I want book'' might have an edit (2-3, ``a book'', ArtOrDet); applying the edit results in ``I want a book''.
Edits are defined (by the annotation guidelines) to be maximally independent, so that each edit can be applied independently of the others.
We denote the examined set of metrics with {\sc METRICS}.

\paragraph{BLEU.}
BLEU \cite{papineni2002bleu} is a reference-based metric that averages the output-reference $n$-gram overlap precision values over different $n$s. 
While commonly used in MT and other text generation tasks \cite{sennrich2017nematus, krishna2017visual, yu2017seqgan}, BLEU was shown to be a problematic metric in 
monolingual translation tasks, in which much of the source sentence should remain unchanged \cite{Xu-EtAl:2016:TACL}.
We use the NLTK implementation of BLEU, using smoothing method 3 by \citet{chen2014systematic}.

 \paragraph{GLEU.}
 GLEU \cite{napoles2015ground} is a reference-based GEC metric inspired by BLEU.
 Recently, it was updated to better address multiple references \cite{napoles2016gleu_update}.
 GLEU rewards $n$-gram overlap of the correction with the reference and penalizes unchanged $n$-grams in the correction that are changed in the reference.

\paragraph{iBLEU.}
iBLEU \cite{sun2012joint_ibleu} was introduced to monolingual translation in order to balance BLEU, by averaging it with the BLEU score of the source and the output. 
This yields a metric that rewards similarity to the source, and not only overlap with the reference:

\begin{small}
\begin{myequation*}
  iBLEU(S,R,O) = \alpha BLEU(O, R) - (1 - \alpha) BLEU(O, S)
\end{myequation*}
\end{small}

\noindent
We set $\alpha=0.8$ as suggested by Sun and Zhou.

\paragraph{$F$-Score} computes the overlap of edits to the source in the reference, and in the output.
  As system edits can be constructed in multiple ways, the standard $M^2$ scorer \cite{dahlmeier2012better} computes
  the set of edits that yields the maximum $F$-score. 
  As $M^2$ requires edits from the source to the reference, and as \maege\ generates new source sentences, we use an established protocol to automatically construct edits from pairs of strings
  \citep{felice-bryant-briscoe:2016:COLING,bryant-felice-briscoe:2017:Long}.
  The protocol was shown to produce similar $M^2$ scores to those produced with manual edits.
  Following common practice, we use the Precision-oriented $F_{0.5}$. 
 
\paragraph{SARI.}
  SARI \cite{Xu-EtAl:2016:TACL} is a reference-based metric proposed for sentence simplification. SARI averages three scores,
  measuring the extent to which $n$-grams are correctly added to the source, deleted from it and retained in it. 
  Where multiple references are present, SARI's score is determined not as the maximum single-reference score, but
  some averaging over them. 
  As this may lead to an unintuitive case, where a correction which is identical to the output gets a score of less than 1,
  we experiment with an additional metric, MAX-SARI,
  which coincides with SARI for a single reference, and computes the maximum single-reference SARI score for multiple-references.
 
\paragraph{Levenshtein Distance.}
  We use the Levenshtein distance \cite{kruskal1983time},
  i.e., the number of character edits needed to convert one string to another,
  between the correction and its closest reference ($MinLD_{O\rightarrow R}$).
  To enrich the discussion, 
  we also report results with a measure of conservatism, $LD_{S\rightarrow O}$, i.e., the Levenshtein distance
  between the correction and the source.
  Both distances are normalized by the number of characters in the second string ($R, O$ respectively).
  In order to convert these distance measures into measures of similarity, we report $1 - \frac{LD(c1, c2)}{len(c1)}$.
  

\paragraph{Grammaticality}
  is a reference-less metric, which uses grammatical error detection tools to assess the grammaticality of GEC system outputs.
  We use LT \cite{milkowski2010developing}, the best performing non-proprietary grammaticality metric \cite{napoles-sakaguchi-tetreault:2016:EMNLP2016}.
  The detection tool at the base of LT can be much improved. Indeed, \citet{napoles-sakaguchi-tetreault:2016:EMNLP2016} reported 
  that the proprietary tool they used detected 15 times more errors than LT. 
  A sentence's score is defined to be $1 - \frac{\#errors}{\#tokens}$. 
  See \cite{asano2017reference,choshen2018usim} for additional reference-less measures, published concurrently with this work.

\paragraph{I-Measure.}
\begin{figure}
	\includegraphics[width=0.9\columnwidth]{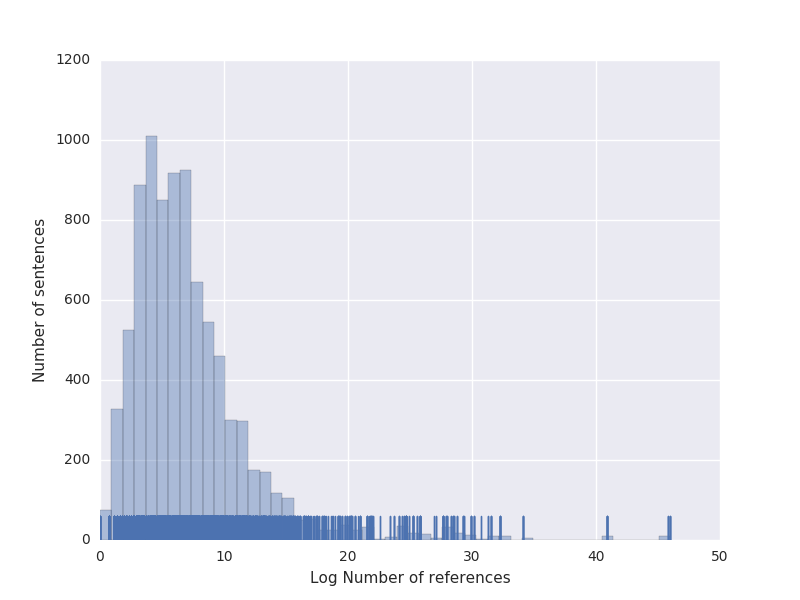}
	\caption{Histogram and rug plot of the log number of references under I-measure assumptions, i.e. overlapping edits alternate as
	valid corrections of the same error. There are billions of ways to combine 8 references on average.\label{fig:im}}
\end{figure}
  I-Measure \cite{felice2015towards} is a weighted accuracy metric over tokens.
  I-measure rank determines whether a correction is better than the source and to what extent. 
  Unlike in this paper, I-measure assumes that every pair of intersecting edits (i.e., edits whose spans of tokens overlap) are alternating,
  and that non-intersecting edits are independent. Consequently, where multiple references are present, it extends the set of
  references, by generating every possible combination of independent edits. As the number of combinations is generally exponential in the number of references,
  the procedure can be severely inefficient.
  Indeed, a sentence in the test set has 3.5 billion references on average, where the median is $512$ (See Figure \ref{fig:im}).
  I-measure can also be run without generating new references, but despite parallelization efforts,
  this version did not terminate after 140 CPU days, while the cumulative CPU time of the rest of the metrics was less than 1.5 days.


\section{Human Ranking Experiments}\label{sec:Human judgments}

Correlation with human rankings (CHR) is the standard methodology for assessing the validity of GEC metrics.
While informative, human rankings are costly to produce, present low inter-rater agreement (shown for MT evaluation in \cite{bojar2011grain, dras2015evaluating}), 
and introduce methodological difficulties that are hard to overcome.
We begin by showing that existing sets of human rankings produce inconsistent results
with respect to the quality of different metrics, and proceed by proposing
an improved protocol for computing this correlation in the future.


There are two existing sets of human rankings for GEC that were compiled concurrently: 
GJG15 by \citet{grundkiewicz2015human}, and NSPT15 by \citet{napoles2015ground}.
Both sets are based on system outputs from the CoNLL 2014 \cite{ng2014conll} shared task,
using sentences from the NUCLE test set.
We compute CHR against each.
System-level correlations are computed by TrueSkill \cite{sakaguchi2014efficient}, which adopts its methodology from 
MT.\footnote{There's a minor problem in the output of the NTHU system: a part of the input is given as sentence 39
and sentence 43 is missing. We corrected it to avoid unduly penalizing NTHU for all the sentences in this range.}

Table \ref{tab:Human judgments} shows CHR with Spearman $\rho$ (Pearson $r$ shows similar trends).
Results on the two datasets diverge considerably, despite their use of the same systems
and corpus (albeit a different sub-set thereof). 
For example, BLEU receives a high positive correlation on GJG15, but a negative one on NSPT15;
GLEU receives a correlation of 0.51 against GJG15 and 0.76 against NSPT15; and $M^2$ ranges 
between 0.4 (GJG15) and 0.7 (NSPT15). In fact, this variance is already apparent in the \emph{published}
correlations of GLEU, e.g., \citet{napoles2015ground} reported a $\rho$ of 0.56 against NSPT15 and \citet{napoles-sakaguchi-tetreault:2016:EMNLP2016} reported a $\rho$ of 0.85 against GJG15.\footnote{The difference between our results
and previously reported ones is probably due to a recent update in GLEU to better tackles multiple references \cite{napoles2016gleu_update}.}
This variance in the metrics' scores is an example 
of the low agreement between human rankings, echoing similar findings in MT \cite{bojar2011grain, lopez2012putting, dras2015evaluating}.

Another source of inconsistency in CHR is that the rankings are relative and sampled, 
so datasets rank different sets of outputs \cite{lopez2012putting}.
For example, if a system is judged against the best systems more often 
then others, it may unjustly receive a lower score.
TrueSkill is the best known practice to tackle such issues \cite{bojar2014findings},
but it produces a probabilistic corpus-level score, which can vary between 
runs \cite{sakaguchi2016reassessing}.\footnote{The standard deviation of the results is about 0.02.}
This makes CHR more difficult to interpret, compared to classic correlation coefficients.

We conclude by proposing a practice for reporting CHR in future work.
First, we combine both sets of human judgments to arrive at the statistically most powerful test. 
Second, we compute the metrics' corpus-level rankings according to the same subset of sentences
used for human rankings. The current practice of allowing metrics to rank systems based
on their output on the entire CoNLL test set (while human rankings are only collected for a sub-set thereof), 
may bias the results due to potential non-uniform system performance on the test set. 
We report CHR according to the proposed protocol in Table \ref{tab:Human judgments} (left column).


\begin{table}[t]
	\centering
	\small
	\singlespacing
	\resizebox{\columnwidth}{!}{
	    \begin{tabular}{@{}l|rl|rl|rl@{}}
        \toprule
		    \com{& \multicolumn{2}{l|}{Combined} & \multicolumn{2}{l|}{NSPT15} & \multicolumn{2}{l}{GJG15} \\ 
		    & $\rho$         & P-val       & $\rho$     & P-val     & $\rho$         & P-val        \\
        \midrule
		    GLEU        & 0.771     & 0.001   & 0.512 & 0.061 & 0.758     & 0.003    \\
		    LT     & 0.692     & 0.006   & 0.358 & 0.208 & 0.615     & 0.025    \\ 
		    $M^2$          & 0.626     & 0.017   & 0.398 & 0.159 & 0.703     & 0.007    \\
		    SARI        & 0.596     & 0.025   & 0.323 & 0.260 & 0.599     & 0.031    \\
		    MAX-SARI    & 0.552     & 0.041   & 0.292 & 0.311 & 0.577     & 0.039    \\
		    
		    $MinLD_{O\rightarrow R}$ & 0.191    & 0.513    & 0.35 & 0.21 & -0.187    & 0.54    \\
		    BLEU        & 0.143     & 0.626   & 0.455 & 0.102 & -0.126    & 0.681    \\
		    iBLEU        & -0.059     & 0.84   & 0.226 & 0.436 & -0.462    & 0.112    \\
        \midrule
        $LD_{S\rightarrow O}$ & -0.481    & 0.081    & -0.178 & 0.543 & -0.505    & 0.078    \\
}
    & \multicolumn{2}{l|}{Combined} & \multicolumn{2}{l|}{GJG15} & \multicolumn{2}{l}{NSPT15} \\ 
    & $\rho$         & P-val       & $\rho$     & Rank     & $\rho$         & Rank        \\
    \midrule
    GLEU        				& 0.771     & 0.001	& 0.512 & 1 & 0.758     & 1    \\
    LT   						& 0.692     & 0.006	& 0.358 & 4 & 0.615     & 3    \\ 
    $M^2$       				& 0.626     & 0.017	& 0.398 & 3 & 0.703     & 2    \\
    SARI        				& 0.596     & 0.025	& 0.323 & 6 & 0.599     & 4    \\
    MAX-SARI	    			& 0.552     & 0.041	& 0.292 & 7 & 0.577     & 5    \\
    $MinLD_{O\rightarrow R}$ 	& 0.191		& 0.513	& 0.350 & 5 	& -0.187    & 7    \\
    BLEU        				& 0.143     & 0.626	& 0.455 & 2 & -0.126    & 6    \\
    iBLEU        				& -0.059    & 0.840	& 0.226 & 8 & -0.462    & 8    \\
    \midrule   
    $LD_{S\rightarrow O}$ 		& -0.481    & 0.081	& -0.178 &  & -0.505    &     \\
\bottomrule
	    \end{tabular}
  }
	\caption{Metrics correlation with human judgments.
	         The {\it Combined} column presents the Spearman correlation coefficient ($\rho$) 
	         according to the combined set of human rankings, with its associated P-value.
	         The GJG15 and NSPT15 columns present the Spearman correlation according to the two sets 
	         of human rankings, as well as the rank of the metric according to this correlation. 
	         Measures are ordered by their rank in the combined human judgments.
	         The discrepancy between the $\rho$ values obtained against GJG15 and NSPT15 demonstrate low inter-rater agreement in human rankings.
	         \label{tab:Human judgments}}
\end{table}


\section{Constructing Lattices of Corrections}\label{sec:maege}

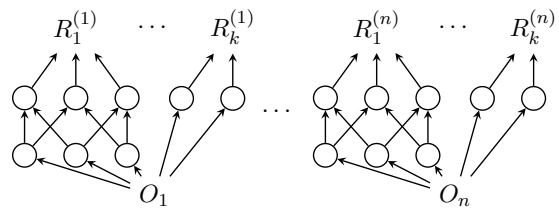
\begin{figure}
  \resizebox{\columnwidth}{!}{
	\centering
		\def\l{-2.3}
		\def\h{2.6}
		\def\rs{1.2}
		\begin{tikzpicture}[ 
		node distance=5mm and 4mm, semithick
		]
		\node at(0, 0) 	(zero)  		     	   	{};
		\node at(\l - \rs, \h) 	(R1)  		     	   	{$R_1^{(1)}$};
		\node at(\l + \rs, \h) 	(Rk)        	 		{$R_k^{(1)}$};
		\path (R1) -- (Rk) node[midway] (dots)  {$\cdots$};
		\node [circle, draw=black]	(midkr) [below=of Rk]    	{};
		\node [circle, draw=black](mid0 2) [below=of R1]    	{};
		\node [circle, draw=black]	(mid0 1) [right=of mid0 2] {};	
		\node [circle, draw=black]	(mid0 3) [left=of mid0 2]  {};
		\node [circle, draw=black]	(bot0 2) [below=of mid0 2] {};
		\node [circle, draw=black]	(bot0 1) [right=of bot0 2] {};	
		\node [circle, draw=black]	(bot0 3) [left=of bot0 2]  {};
		\node [circle, draw=black]	(midkl) 	[left=of midkr] {};
		\node at (\l ,0) 		(S)      				   {$O_1$};
		\path[-stealth]
		(S) edge  (bot0 1)
		(S) edge  (bot0 2)
		(S) edge  (bot0 3)
		(bot0 1) edge  (mid0 1)
		(bot0 1) edge  (mid0 2)
		(bot0 2) edge  (mid0 1)
		(bot0 2) edge  (mid0 3)
		(bot0 3) edge  (mid0 2)
		(bot0 3) edge  (mid0 3)
		(mid0 1) edge  (R1)
		(mid0 2) edge  (R1)
		(mid0 3) edge  (R1)
		(midkr) edge (Rk)
		(midkl) edge (Rk)
		(S) edge (midkl)
		(S) edge (midkr);
		\node at(-\l - \rs, \h) 	(nR1)  		     	   	{$R_1^{(n)}$};
		\node at(-\l + \rs, \h) 	(nRk)        	 		{$R_k^{(n)}$};
		\node at (-0.4, \h/2)  	(middots)  				{$\cdots$};
		\path (nR1) -- (nRk) node[midway] (ndots)  {$\cdots$};
		\node [circle, draw=black]	(nmidkr) [below=of nRk]    	{};
		\node [circle, draw=black]	(nmid0 2) [below=of nR1]    	{};
		\node [circle, draw=black]	(nmid0 1) [right=of nmid0 2] {};	
		\node [circle, draw=black]	(nmid0 3) [left=of nmid0 2]  {};
		\node [circle, draw=black]	(nbot0 2) [below=of nmid0 2] {};
		\node [circle, draw=black]	(nbot0 1) [right=of nbot0 2] {};	
		\node [circle, draw=black]	(nbot0 3) [left=of nbot0 2]  {};
		\node [circle, draw=black]	(nmidkl) 	[left=of nmidkr] {};
		\node at (-\l ,0) 		(nS)      				   {$O_n$};
		\path[-stealth]
		(nS) edge  (nbot0 1)
		(nS) edge  (nbot0 2)
		(nS) edge  (nbot0 3)
		(nbot0 1) edge  (nmid0 1)
		(nbot0 1) edge  (nmid0 2)
		(nbot0 2) edge  (nmid0 1)
		(nbot0 2) edge  (nmid0 3)
		(nbot0 3) edge  (nmid0 2)
		(nbot0 3) edge  (nmid0 3)
		(nmid0 1) edge  (nR1)
		(nmid0 2) edge  (nR1)
		(nmid0 3) edge  (nR1)
		(nmidkr) edge (nRk)
		(nmidkl) edge (nRk)
		(nS) edge (nmidkl)
		(nS) edge (nmidkr);
		\end{tikzpicture}
    }
		\caption{An illustration of the generated corrections lattices. The $O_i$s are the original sentences, directed edges represent an application of an edit and $R_j^{(i)}$ 
		         is the $j$-th perfect correction of $O_i$ (i.e., the perfect correction that result from applying all the edits of the $j$-th annotation of $O_i$).
            \label{fig:lattice}}
\end{figure}

%
In the following
sections we present \maege\, an alternative methodology to CHR, which uses human corrections to induce
more reliable and scalable rankings to compare metrics against.
We begin our presentation by detailing the method  \maege\ uses to generate source-correction
pairs and a partial order between them.
\maege\ operates by using a corpus with gold annotation, given as edits, 
to generate lattices of corrections, each defined by a sub-set of the edits.
Within the lattice, every pair of sentences can be regarded as a potential source
and a potential output.
We create sentence chains, in an increasing order of quality, taking a 
source sentence and applying edits in some order one after the other (see Figure \ref{fig:lattice} and \ref{fig:example_chain}).

Formally, for each sentence $s$ in the corpus and each annotation $a$, 
we have a set of typed edits $edits(s,a)=\{e_{s,a}^{(1)},\ldots,e_{s,a}^{(n_{s,a})}\}$ of size $n_{s,a}$.
We call $2^{edits(s,a)}$ the {\it corrections lattice}, and denote it with $E_{s,a}$.
We call, $s$, the correction corresponding to $\emptyset$ the {\it original}.
We define a partial order relation between $x,y \in E_{s,a}$ such that $x < y$ if $x \subset y$.
This order relation is assumed to be the gold standard ranking between the corrections.

\begin{figure}
	\resizebox{\columnwidth}{!}{
		\centering
		\def\l{-2.3}
		\def\h{2.6}
		\def\rs{1.2}
		\begin{tikzpicture}[ 
		node distance=5mm and 4mm, semithick
		]
		\node at (\l ,0) 		(S)      				   {
			\begin{tabular}{cc}
			Social media makes our life patten so fast  & \\
			and left us less time to think about our life. & \\
			\end{tabular}
		};
		\node (S1) 	[above=of S] {\begin{tabular}{cc}
			Social media makes our life patten so fast  & \\
			and \textbf{leave} us less time to think about our life. & \\
			\end{tabular}
		};
		\node (S2) 	[above=of S1] {\begin{tabular}{cc}
			Social media \textbf{make} our life patten so fast  & \\
			and leave us less time to think about our life. & \\
			\end{tabular}
		};
		\node (S3) 	[above=of S2] {\begin{tabular}{cc}
			Social media make our \textbf{pace of life} so fast  & \\
			and leave us less time to think about our life. & \\
			\end{tabular}
		};
		\path[-stealth]
		(S) edge  node[right] {\sout{left} leave} (S1)
		(S1) edge  node[right] {\sout{makes} make} (S2)
		(S2) edge  node[right] {\sout{life patten} pace of life} (S3);
		\end{tikzpicture}
	}
	\caption{An example chain from a corrections lattice -- each sentence is the result of applying
		a single edit to the sentence below it. The top sentence is a perfect correction,
		while the bottom is the original.\label{fig:example_chain}}
\end{figure}

For our experiments, we use the NUCLE test data \cite{ng2014conll}. Each sentence is paired with two annotations.
The other eight available references, produced by \citet{bryant2015far}, are used as references for the 
reference-based metrics. Denote the set of references for $s$ with $R_s$.

Sentences which require no correction according to at least one of the two annotations are discarded.
In 26 cases where two edit spans intersect in the same annotation (out of a total of about 40K edits),
the edits are manually merged or split.

\begin{table*}[]	
	\centering
	\small
	\singlespacing
	\begin{tabular}{@{}lrr|rl|rl@{}}
		\toprule
		& \multicolumn{2}{c|}{Corpus-level} & \multicolumn{4}{c}{Sentence-level} \\ 
		& \multicolumn{1}{c}{$\rho$}   & P-val          & \multicolumn{1}{c}{$r$}      & P-val           & \multicolumn{1}{c}{$\tau$}         & P-val            \\
  	\midrule
		iBLEU                     & 0.418       & 0.200        & \textbf{0.230}          & $\dagger$          & 0.050              & $\dagger$           \\
		$M^2$       & 0.060 & 0.853      & -0.025& 0.024       & 0.213              & $\dagger$         \\
		LT                        & \textbf{0.973} & $\dagger$         & 0.167       & $\dagger$          & \textbf{0.222} &  $\dagger$           \\
		BLEU                      & 0.564 & 0.071      & 0.214 & $\dagger$          & 0.111              & $\dagger$        \\
		$MinLD_{O\rightarrow R}$ & -0.867 &  $\dagger$         & 0.011 & 0.327       & -0.183            & $\dagger$   \\
		GLEU                      & 0.736& 0.001      & 0.189 &  $\dagger$          & -0.028             & $\dagger$  \\
		MAX-SARI                  & -0.809 &  0.003      & 0.027 &  0.015       & -0.070              & $\dagger$   \\
		SARI                      & -0.545 & 0.080       & 0.061 &  $\dagger$          & -0.039             & $\dagger$   \\ \midrule 
		$LD_{S\rightarrow O}$    & -0.118 &  0.729      & 0.109 &  $\dagger$          & 0.094              & $\dagger$   \\
		\bottomrule
	\end{tabular}
		\caption{Corpus-level Spearman $\rho$, sentence-level Pearson $r$ and Kendall $\tau$ with the metrics (left). $\dagger$ represents P-value $< 0.001$. LT correlates best at the corpus level and has the highest sentence-level $\tau$, 
		while iBLEU has the highest sentence-level $r$. \label{tab:correlations}}
\end{table*}

\section{Corpus-level Analysis}\label{sec:corpus}
\begin{figure}
	\includegraphics[width=0.9\columnwidth]{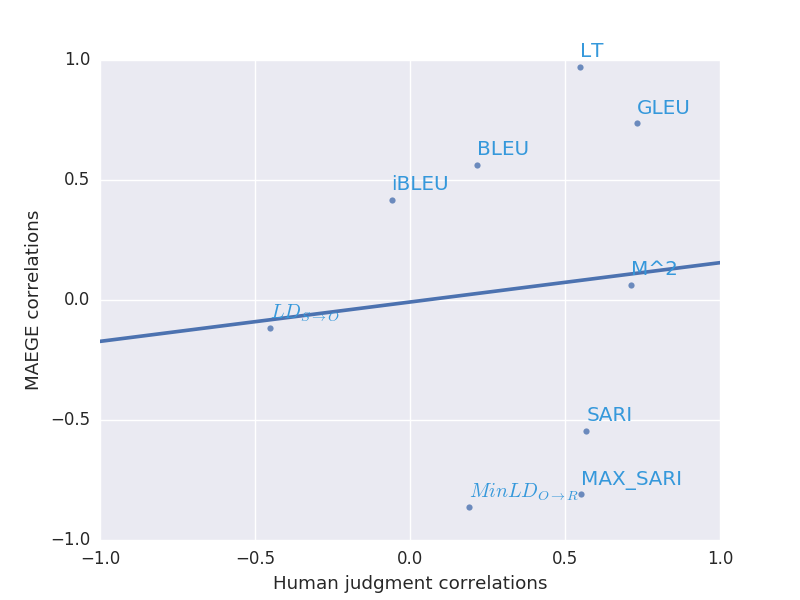}
	\caption{A scatter plot of the corpus-level correlation of metrics according to the different methodologies.
	         The x-axis corresponds to the correlation according to human rankings ({\it Combined} setting), and the y-axis
	         corresponds to the correlation according to \maege. While some get similar correlation (e.g., GLEU), other metrics change drastically (e.g., SARI).
	         \label{fig:correlations}}
\end{figure}

We conduct a corpus-level analysis, namely testing the ability of metrics to determine which corpus of corrections 
is of better quality. In practice, this procedure is used to rank systems based on their outputs on the test corpus.

In order to compile corpora corresponding to systems of different quality levels, we define several corpus models, each 
applying a different expected number of edits to the original.
Models are denoted with the expected number of edits they apply to the original 
which is a positive number $M \in \mathbb{R^+}$.
Given a corpus model $M$, we generate a corpus of corrections
by traversing the original sentences, and for each sentence $s$ uniformly 
sample an annotation $a$ (i.e., a set of edits that results in a perfect correction), 
and the number of edits applied $n_{edits}$, which is sampled from a clipped binomial probability with mean $M$ and variance 0.9.
Given $n_{edits}$, we uniformly sample from the lattice $E_{s,a}$ a sub-set of edits of size $n_{edits}$, 
and apply this set of edits to $s$.
The corpus of $M=0$ is the set of originals.

The corpus of source sentences, against which all other corpora are compared,
is sampled by traversing the original sentences, and for each
sentence $s$, uniformly sample an annotation $a$, and given $s,a$, uniformly sample a sentence
from $E_{s,a}$.

Given a metric $m \in$ {\sc METRICS}, we compute its score for each sampled corpus.
Where corpus-level scores are not defined by the metrics themselves, we use the average sentence score instead.
We compare the rankings induced by the scores of $m$ and the ranking of systems according to their corpus model (i.e.,
systems that have a higher $M$ should be ranked higher), and report the correlation between these rankings.

\subsection{Experiments}
\paragraph{Setup.}
For each model, we sample one correction per NUCLE sentence, noting that it is possible to reduce the variance of the metrics' 
corpus-level scores by sampling more. 
Corpus models of integer values between 0 and 10 are taken.
We report Spearman $\rho$, commonly used for system-level rankings \cite{bojar2017findings}.\footnote{Using Pearson correlation shows similar trends.}

\paragraph{Results.}
  Results, presented in Table \ref{tab:correlations} (left part), shows that LT correlates best with the rankings induced by \maege,
  where GLEU is second. $M^2$'s correlation is only 0.06. We note that the LT
  requires a complementary metric to penalize grammatical outputs that diverge in meaning from the source \cite{napoles-sakaguchi-tetreault:2016:EMNLP2016}. See \S\ref{sec:discussion}.
 
  Comparing the metrics' quality in corpus-level evaluation with their quality according to CHR (\S\ref{sec:Human judgments}), we
  find they are often at odds. Figure \ref{fig:correlations} plots the Spearman correlation of the different metrics according 
  to the two validation methodologies,
  showing correlations are slightly correlated, but disagreements as to metric quality are frequent and substantial (e.g., with
  iBLEU or SARI).



\section{Sentence-level Analysis}\label{sec:sentence}

\begin{figure}
	\includegraphics[width=0.9\columnwidth]{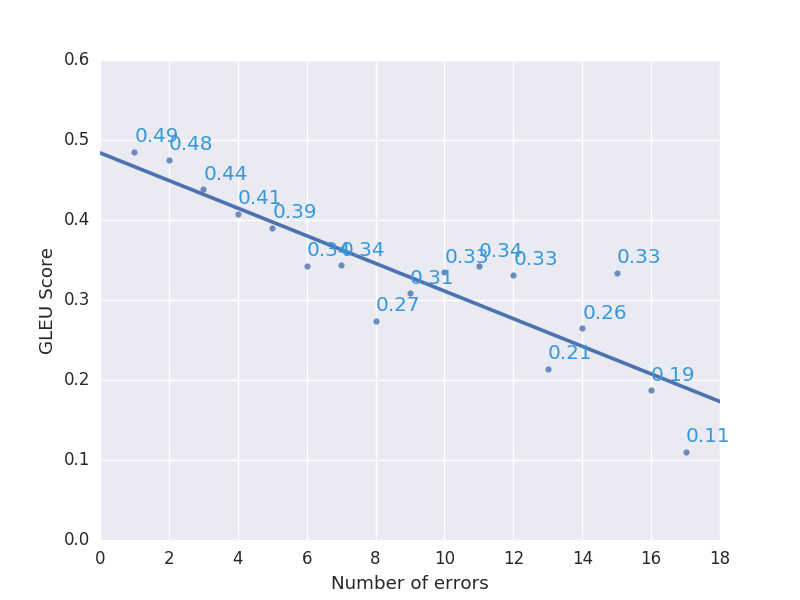}
	\caption{Average GLEU score of originals (y-axis), plotted against the number of errors they contain (x-axis). 
	Their substantial correlation indicates that GLEU is globally reliable.\label{fig:gleu}}
\end{figure}


We proceed by presenting a method for assessing the correlation between metric-induced scores of corrections of the same sentence, 
and the scores given to these corrections by \maege.
Given a sentence $s$ and an annotation $a$, we sample a random permutation over the edits in $edits(s,a)$.
We denote the permutation with $\sigma \in S_{n_{s,a}}$, where $S_{n_{s,a}}$ is the permutation group over $\{1,\cdots,n_{s,a}\}$.
Given $\sigma$, we define a monotonic chain in $E_{i,j}$ as:

\begin{small}
\begin{align*}
chain(s,a,\sigma) = \Big( \emptyset < \{e_{s,a}^{(\sigma(1))}\} < \{e_{s,a}^{(\sigma(1))},e_{s,a}^{(\sigma(2))}\} < \\ 
 \ldots < edits(s,a) \Big)
\end{align*}
\end{small}
 
\noindent
For each chain, we uniformly sample one of its elements, mark it as the source, and denote it with $src$.
In order to generate a set of chains, \maege\ traverses the original sentences and annotations, and for each
sentence-annotation pair, uniformly samples $n_{ch}$ chains without repetition.
It then uniformly samples a source sentence from each chain.
If the number of chains in $E_{s,a}$ is smaller than $n_{ch}$, \maege\ selects all the chains.

Given a metric $m \in$ {\sc METRICS}, we compute its score for every correction in each sampled chain against the sampled
source and available references.
We compute the sentence-level correlation of the rankings induced by the scores of $m$ and the rankings induced by $<$. 
For computing rank correlation (such as Spearman $\rho$ or Kendall $\tau$), such a relative ranking is sufficient.

We report Kendall $\tau$, which is only sensitive to the relative ranking of correction pairs within the same chain.
Kendall is minimalistic in its assumptions, as it does not require numerical scores, 
but only assuming that $<$ is well-motivated, i.e., 
that applying a set of valid edits is better in quality than applying only a subset of it.

As $<$ is a partial order, and as Kendall $\tau$ is standardly defined over total orders, some modification
is required. $\tau$ is a function of the number of compared pairs and of discongruent pairs (ordered differently in the compared rankings):
  
  \begin{myequation*}
     \tau = 1-\frac{2 \left|\text{discongruent pairs}\right|}{\left|\text{all pairs}\right|}.
  \end{myequation*}

 To compute these
quantities, we extract all unique pairs of corrections that can be compared with $<$ 
(i.e., one applies a sub-set of the edits of the other), and count the number of discongruent ones
between the metric's ranking and $<$.
Significance is modified accordingly.\footnote{Code can be found in \url{https://github.com/borgr/EoE}}
Spearman $\rho$ is less applicable in this setting, as it compares total orders whereas here we compare partial orders.

To compute linear correlation with Pearson $r$, we make the simplifying assumption that all edits contribute 
equally to the overall quality. Specifically, we assume that a perfect correction (i.e., the top of a chain) receives a score of 1. 
Each original sentence $s$ (the bottom of a chain), for which there exists annotations $a_1,\ldots,a_n$, receives a score of

\begin{myequation*}
1 - \min_i \frac{\left|edits(s,a_i)\right|}{\left|tokens(s)\right|}.
\end{myequation*}

The scores of partial (non-perfect) corrections in each chain are linearly spaced between the score of the perfect correction and that of the original.
This scoring system is well-defined, as a partial correction receives the same score according to all chains it is in,
as all paths between a partial correction and the original have the same length.

\npdecimalsign{.}
\nprounddigits{3}
\begin{table*}[]
	\centering
	\small
	\singlespacing
	\resizebox{\textwidth}{!}{%
		\begin{tabular}{@{}l|n{5}{2}n{5}{2}n{5}{2}n{5}{2}n{5}{2}n{5}{2}n{5}{2}n{5}{2}|n{5}{2}@{}}

			Type & 
			\multicolumn{1}{c}{\quad\quad iBLEU}         &
			 \multicolumn{1}{c}{\hspace{.8cm} $M^2$}         & \multicolumn{1}{c}{\quad \quad LT}            &
			\multicolumn{1}{c}{\quad\quad BLEU}          &
			\multicolumn{1}{c}{\quad$MinLD_{O\rightarrow R}$} &
			 \multicolumn{1}{c}{\quad\quad GLEU} &
			  \multicolumn{1}{c}{\quad MAX-SARI}     &			\multicolumn{1}{c}{\quad\quad SARI}          & 
			  \multicolumn{1}{|c}{\hspace{0.4cm} $LD_{S\rightarrow O}$}\\
			\toprule
			WOinc    & 0.0161721163  & -0.0004996091       & -0.001778213  & -0.0048166379 & -0.0258943146             & -0.0507862    & -0.0751595672 & -0.0456813869 & 0.0631577912            \\
			Nn       & 0.0327965568  & -0.0008998388       & 0.0037272707  & 0.0289575016  & -0.0065248033             & 0.0245424843  & 0.0432113275  & 0.0370516449& 0.0170710429             \\
			Npos     & -0.0010871796 & 0.0005341006        & -0.0038950919 & -0.0111696302 & -0.0074523748             & -0.0301756852 & -0.0225210019 & -0.0086233951 & 0.013804637            \\
			Sfrag    & -0.0249352396 & -0.0032518741       & -0.0004917122 & -0.0669403346 & -0.068040136              & -0.1427227333 & -0.1773769361 & -0.1424439565  & 0.0763608244           \\
			Wtone    & -0.0126634327 & -0.0022971711       & -0.0080440308 & -0.0241027591  & -0.0208240017   & -0.0262214643 & -0.0858282181 & -0.0548215344 & 0.0182494363           \\
			Srun     & -0.0265907221 & -0.0044574191       & -0.0037131958 & -0.0477532796 & -0.0142628907             & -0.0781482667 & -0.0388065121 & -0.0296012966  & 0.0200227165          \\
			ArtOrDet & 0.0281699304  & -0.0007073084       & 0.0008667759  & 0.0186567544   & -0.0063117856             & -0.0032393651 & -0.0221566977 & -0.0034484087  & 0.0241740972         \\
			Vt       & 0.0541764029  & -0.0009333635       & 0.0054548203  & 0.0455630711  & -0.0021204362             & 0.0107856364  & 0.0027273904  & 0.0179214891    & 0.0248060407         \\
			Wa       & 0.0407118479  & -0.0023793338       & -0.0020703934 & -0.0131842894       & 0.0055304173              & -0.027972     & -0.0726840774 & -0.089545413  & 0.0711466165     \\
			Wform    & 0.0489861468  & -0.0008944842       & 0.0018691842  & 0.0439472359      & -0.0028498321             & 0.0099072708  & 0.0038378171  & 0.0204599543    & 0.0223781805     \\
			WOadv    & 0.0074041246  & 0.000282521         & 0.0087324911  & 0.0106707439    & 0.0116642791              & 0.0058221667  & 0.087512316   & 0.054496206    & -0.0144637807       \\
			V0       & 0.0151512429  & -0.0010500731       & 0.0186269948  & 0.0046802534    & -0.0030244751             & -0.0062995541 & -0.0102988816 & -0.0035749641  & 0.0153787473        \\
			Trans    & -0.0111309022 & 0.0003340485        & 0.0051963392  & -0.0218580424      & -0.0286869045             & -0.0308815208 & -0.0193275095 & -0.0041098452  & 0.0130306099     \\
			Pform    & 0.0209573705  & -0.0009266493       & 0.0028225381  & 0.0114303295    & -0.0117086462             & -0.0186641711 & -0.0030750096 & 0.0048583803    & 0.0297980494       \\
			Smod     & -0.0518852521 & 0.0011929502        & 0.0039267702  & -0.0925426616      & -0.0716341098             & -0.1255171667 & -0.0622964625 & -0.042808857  & 0.0554564541      \\
			Ssub     & -0.0051550943 & 0.0002343963        & -0.01125337   & -0.0242830137    & -0.0270453149             & -0.0524124382 & -0.0722228042 & -0.0383203722 & 0.0264501872        \\
			Wci      & -0.0076208381 & -0.0006346588       & 0.0039835564  & -0.0219914024 & -0.0290446011             & -0.0446959045 & -0.0493262458 & -0.0317218431  & 0.0168266205          \\
			Vm       & -0.0067142152 & -0.0008082089       & -0.000622482  & -0.0287459854     & -0.026519626              & -0.0746720109 & -0.0697491259 & -0.0594857692 & 0.0304513937       \\
			Pref     & -0.0025982978 & -0.0012535072       & 0.0019275275  & -0.0150291655       & -0.0217696725             & -0.0453845872 & -0.0478963141 & -0.0345900614  & 0.0178315174    \\
			Mec      & 0.0123959188  & 0.0010275384        & 0.0135232033  & 0.0035595716          & -0.0126376151             & -0.0137984651 & 0.0001550709  & 0.0020035928   & 0.0178510203  \\
			Vform    & 0.0434589143  & -0.0008637425       & 0.0055945448  & 0.0441764811        & 0.00003                  & 0.029850736   & 0.0331057535  & 0.0425124575    & 0.0127873423   \\
			Prep     & 0.0183305492  & -0.0004522389       & 0.0037969069  & 0.0110614689          & -0.0077209281             & -0.0012861814 & -0.0100668054 & 0.0047177935  & 0.0144461345   \\
			Um       & -0.0378528061 & -0.0011222254 &        -0.0074394698 & -0.0434615047         & -0.1000975877             & -0.0373438852 & -0.046468427  & -0.0320366867 & 0.0085987819   \\
			Others   & -0.048246677  & -0.00003           & 0.0070635231  & -0.0627705199        & -0.0542578439             & -0.0597926559 & -0.0396035441 & -0.0240197112  & -0.0003407076  \\
			Rloc-    & 0.0039141544  & -0.0008522991       & -0.0038089926 & -0.0062519764         & -0.026971383              & -0.0231590476 & -0.0281521328 & -0.0187993787 & 0.0217493729   \\
			Spar     & 0.0410994907  & 0.0005999215        & 0.003315884   & 0.0347920698         & -0.0120114432             & -0.0031445    & 0.0077243351  & 0.0257906686   & 0.0238573056   \\
			SVA      & 0.0451797841  & -0.0006982543       & -0.0009767204 & 0.0371730455          & -0.0052847683             & -0.0015608889 & 0.0117283336  & 0.014534758   & 0.0213210787  
		\end{tabular}%
	}
	\caption{Average change in metric score by metric and edit types ($\Delta_{m,t}$; see text). Rows correspond to edit types (abbreviations in \citet{dahlmeier2013building}); columns correspond to metrics. 
	Some edit types are consistently penalized.
	}
	\label{tab:types}
\end{table*}

\subsection{Experiments}
\paragraph{Setup.}
  We experiment with $n_{ch}=1$,
  yielding 7936 sentences in 1312 chains (same as the number of original sentences in the NUCLE test set).
  We report the Pearson correlation over the scores of all sentences in all chains ($r$), and 
  Kendall $\tau$ over all pairs of corrections within the same chain.
    
\paragraph{Results.}
  Results are presented in Table \ref{tab:correlations} (right part). 
  No metric scores very high, neither according to Pearson $r$ nor according to Kendall $\tau$.
  iBLEU correlates best with $<$ according to $r$, obtaining a correlation of 0.23, whereas LT fares
  best according to $\tau$, obtaining 0.222.
  
  Results show a discrepancy between the low corpus-level and sentence-level $r$ correlations of $M^2$ and its high sentence-level $\tau$.
  It seems that although $M^2$ orders pairs of corrections well, its scores are
  not a linear function of \maege's scores. This may be due to $M^2$'s assignment of the minimal possible score to 
  the source, regardless of its quality.
  $M^2$ thus seems to predict well the relative quality of corrections of the same sentence,
  but to be less effective in yielding a globally coherent score (cf. \citet{felice2015towards}).
  
  GLEU shows the inverse behaviour, failing to correctly order pairs of corrections of the same sentence,
  while managing to produce globally coherent scores. 
  We test this hypothesis by computing the average difference in GLEU score between all pairs in the sampled chains,
  and find it to be slightly negative (-0.00025), which is in line with GLEU's small negative $\tau$. 
  On the other hand, plotting the GLEU scores of the originals grouped by the number of errors they contain,
  we find they correlate well (Figure~\ref{fig:gleu}), indicating that GLEU performs well in comparing 
  the quality of corrections of different sentences. Four sentences with considerably 
  more errors than the others were considered outliers and removed.

\section{Metric Sensitivity by Error Type} \label{sec:types}
\maege's lattice can be used to analyze 
how the examined metrics reward corrections of errors of different types.
For each edit type $t$, we denote with $S_t$ the set of correction pairs from the lattice that only differ
in an edit of type $t$.
For each such pair $(c,c')$ and for each metric $m$, we compute the difference in the score assigned by $m$ to $c$ and $c'$.
The average difference is denoted with $\Delta_{m,t}$. 

\begin{myequation*}
  \Delta_{m,t} = \frac{1}{|S_{t}|} \hspace{-.1cm}\sum_{(c,c') \in S_{t}} \hspace{-.2cm} \big[ m(src,c,R) - m(src,c',R) \big]
\end{myequation*}


\noindent
$R$ is the corresponding reference set. 
A negative (positive) $\Delta_{m,t}$ indicates that $m$ penalizes (awards) valid corrections of type $t$.

\subsection{Experiments}
 
\paragraph{Setup.}
We sample chains using the same sampling method as in \S\ref{sec:sentence},
and uniformly sample a source from each chain. For each edit type $t$, we detect all pairs
of corrections in the sampled chains that only differ in an edit of type $t$, 
and use them to compute $\Delta_{m,t}$.
We use the set of 27 edit types given in the NUCLE corpus.


\paragraph{Results.}
Table \ref{tab:types} presents the results, showing that under all metrics, some edits types are penalized and others rewarded. 
iBLEU and LT penalize the least edit types, and 
GLEU penalizes the most, providing another perspective on GLEU's negative Kendall $\tau$ (\S\ref{sec:sentence}).
Certain types are penalized by almost all metrics. One such type is Vm, wrong verb modality (e.g., ``as they [$\emptyset \leadsto$ may] not want to know''). Another such type is Npos, a problem in noun possessive (e.g., ``their [facebook's $\leadsto$ Facebook] page'').
Other types, such as Mec, mechanical (e.g., ``[real-life $\leadsto$ real life]''),
and V0, missing verb (e.g., ``'Privacy', this is the word that [$\emptyset \leadsto$ is] popular''), are often rewarded by the metrics.

In general, the tendency of reference-based metrics (the vast majority of GEC metrics) to penalize edits
of various types suggests that many edit types are under-represented in available reference sets.
Automatic evaluation of systems that perform these edit types may, therefore, be unreliable. 
Moreover, not addressing these biases in the metrics may hinder progress in GEC.
Indeed, $M^2$ and GLEU, two of the most commonly used metrics, only award a small sub-set of edit types,
thus offering no incentive for systems to improve performance on such types.\footnote{$LD_{S\rightarrow O}$
  tends to award valid corrections of almost all types. 
	As source sentences are randomized across chains, this indicates that on average, corrections with more applied edits  
	tend to be more similar to comparable corrections on the lattice.
	This is also reflected by the slightly positive sentence-level correlation of $LD_{S\rightarrow O}$ (\S\ref{sec:sentence}).}





\begin{table*}[]
	\centering
	\small
	\singlespacing
	\begin{tabular}{@{}lll|ll|ll@{}}
		\toprule
		& \multicolumn{2}{c|}{Corpus-level} & \multicolumn{4}{c}{Sentence-level} \\ 
		& $\rho$        & P-val          & $r$        & P-val           & $\tau$        & P-val            \\
		\midrule
		iBLEU                     &  -0.872 (0.418)        & $\dagger$        & \ 0.235 (0.230) &     $\dagger$               & \ 0.053 (0.050)             & $\dagger$            \\
		$M^2$       & \ 0.882 (0.060)        & $\dagger$      & -0.014 (-0.025)       &  0.223   &  \ 0.223 (0.213)              & $\dagger$            \\
		LT                        & \ 0.836 (0.973)        & 0.001          & \ 0.175 (0.167)         & 0.019           & \ 0.184 (0.222)           & $\dagger$            \\
		BLEU                      &  \ 0.845 (0.564)       & 0.001      & \ 0.217 (0.214)       & $\dagger$           & \ 0.115 (0.111)              & $\dagger$         \\
		$MinLD_{O\rightarrow R}$ & -0.909 (-0.867)      & $\dagger$          & \ 0.022 (0.011) &          $\dagger$       & -0.180 (-0.183)            & $\dagger$           \\
		GLEU                      & \ 0.945 (0.736)         & $\dagger$      & \ 0.208 (0.189) &          $\dagger$           & \ 0.003 (-0.028)             & $\dagger$          \\
		MAX-SARI                  & \ 0.772 (-0.809)       & 0.005      & \ 0.053 (0.027) &          $\dagger$       & \ 0.004 (-0.070)              & 0.6           \\
		SARI                      & \ 0.800 (-0.545)       & 0.003       & \ 0.084 (0.061) &          $\dagger$           & \ 0.022 (-0.039)             & 0.001           \\ \midrule
		$LD_{S\rightarrow O}$    &  -0.972 (-0.118)       & $\dagger$      & \ 0.025 (0.109) &          0.027           & \ 0.070 (0.094)              & $\dagger$            \\ \bottomrule
	\end{tabular}
	\caption{Corpus-level Spearman $\rho$, sentence-level Pearson $r$ and Kendall $\tau$ correlations using 
		origin as the source with the various metrics (left). Correlations using a random source are found in parenthesis. $\dagger$ represents $P-value < 0.001$. 
		LT is the best corpus correlated, and has the best $\tau$ while iBLEU has the best $r$ \label{tab:origin_correlations}}
\end{table*}

\section{Discussion}\label{sec:discussion}
We revisit the argument that using system outputs to perform metric validation poses a methodological difficulty.
Indeed, as GEC systems are developed, trained and tested using available metrics, and as metrics tend to reward some 
correction types and penalize others (\S\ref{sec:types}), it is possible that GEC development adjusts
to the metrics, and neglects some error types.
Resulting tendencies in GEC systems would then yield biased sets of outputs for human rankings,
which in turn would result in biases in the validation process.

To make this concrete, GEC systems are often precision-oriented: trained to prefer not to correct than to invalidly correct.
Indeed, \citet{choshen2018conservatism} show that modern systems tend to be highly conservative, often
performing an order of magnitude fewer changes to the source than references do. 
Validating metrics on their ability to rank conservative system outputs (as is de facto the common practice) may produce a different 
picture of metric quality than when considering a more inclusive set of corrections.

We use \maege\ to mimic a setting of ranking against precision-oriented outputs. To do so, we perform corpus-level and sentence-level analyses,
but instead of randomly sampling a source, we invariably take the original sentence as the source. We thereby create a setting
where all edits applied are valid (but not all valid edits are applied).


Comparing the results to the regular \maege\ correlation (Table \ref{tab:origin_correlations}), 
we find that $LT$ remains reliable, 
while $M^2$, that assumes the source receives the worst possible score, gains from this unbalanced setting. 
iBLEU drops, suggesting it may need to be retuned to this setting and give less weight to $BLEU(O,S)$, 
thus becoming more like BLEU and GLEU. 
The most drastic change we see is in SARI and MAX-SARI, which flip their sign and present strong performance.
Interestingly, the metrics that benefit from this precision-oriented setting in the corpus-level
are the same metrics that perform better according to CHR than to \maege\ (Figure \ref{fig:correlations}).
This indicates the different trends produced by \maege\ and CHR, may result from the latter's use of precision-oriented outputs.

\paragraph{Drawbacks.}
Like any methodology \maege\ has its simplifying assumptions and drawbacks; we wish to make them explicit.
First, any biases introduced in the generation of the test corpus are inherited by \maege\ (e.g., that edits are contiguous and independent of each other). 
Second, \maege\ does not include  errors that a human will not perform but machines might, e.g., significantly altering the meaning of the source.
This partially explains why LT, which measures grammaticality but not meaning preservation, excels in our experiments.
Third, \maege's scoring system (\S\ref{sec:sentence}) assumes that all errors damage the score equally.
While this assumption is made by GEC metrics, we believe it should be refined in future work by collecting user information.




\section{Conclusion}

In this paper, we show how to leverage existing annotation in GEC for performing validation reliably.
We propose a new automatic methodology, \maege,
which overcomes many of the shortcomings of the existing methodology.
Experiments with \maege\ reveal a different picture of metric quality than previously reported.
Our analysis suggests that differences in observed metric quality are partly due to system outputs 
sharing consistent tendencies, notably their tendency to under-predict corrections.
As existing methodology ranks system outputs, these shared tendencies bias the validation process.
The difficulties in basing validation on system outputs may be applicable to other text-to-text generation tasks,
a question we will explore in future work.



\section*{Acknowledgments}

This work was supported by the Israel Science Foundation (grant No. 929/17),
and by the HUJI Cyber Security Research Center in conjunction with the Israel
National Cyber Bureau in the Prime Minister's Office.
We thank Joel Tetreault and Courtney Napoles for helpful feedback and inspiring conversations and Or Zuk for the help with the statistics.

\bibliographystyle{acl_natbib}
\bibliography{related_work}

\begin{thebibliography}{36}
\expandafter\ifx\csname natexlab\endcsname\relax\def\natexlab#1{#1}\fi

\bibitem[{Asano et~al.(2017)Asano, Mizumoto, and Inui}]{asano2017reference}
Hiroki Asano, Tomoya Mizumoto, and Kentaro Inui. 2017.
\newblock Reference-based metrics can be replaced with reference-less metrics
  in evaluating grammatical error correction systems.
\newblock In \emph{Proceedings of the Eighth International Joint Conference on
  Natural Language Processing (Volume 2: Short Papers)}, volume~2, pages
  343--348.

\bibitem[{Birch et~al.(2016)Birch, Abend, Bojar, and Haddow}]{birch2016hume}
Alexandra Birch, Omri Abend, Ond\v{r}ej Bojar, and Barry Haddow. 2016.
\newblock \href {https://aclweb.org/anthology/D16-1134} {Hume: Human ucca-based
  evaluation of machine translation}.
\newblock In \emph{Proceedings of the 2016 Conference on Empirical Methods in
  Natural Language Processing}, pages 1264--1274.

\bibitem[{Bojar et~al.(2014)Bojar, Buck, Federmann, Haddow, Koehn, Leveling,
  Monz, Pecina, Post, Saint-Amand et~al.}]{bojar2014findings}
Ondrej Bojar, Christian Buck, Christian Federmann, Barry Haddow, Philipp Koehn,
  Johannes Leveling, Christof Monz, Pavel Pecina, Matt Post, Herve Saint-Amand,
  et~al. 2014.
\newblock Findings of the 2014 workshop on statistical machine translation.
\newblock In \emph{Proceedings of the ninth workshop on statistical machine
  translation}, pages 12--58.

\bibitem[{Bojar et~al.(2017)Bojar, Chatterjee, Federmann, Graham, Haddow,
  Huang, Huck, Koehn, Liu, Logacheva et~al.}]{bojar2017findings}
Ond{\v{r}}ej Bojar, Rajen Chatterjee, Christian Federmann, Yvette Graham, Barry
  Haddow, Shujian Huang, Matthias Huck, Philipp Koehn, Qun Liu, Varvara
  Logacheva, et~al. 2017.
\newblock Findings of the 2017 conference on machine translation (wmt17).
\newblock In \emph{Proceedings of the Second Conference on Machine
  Translation}, pages 169--214.

\bibitem[{Bojar et~al.(2011)Bojar, Ercegov{\v{c}}evi{\'c}, Popel, and
  Zaidan}]{bojar2011grain}
Ond{\v{r}}ej Bojar, Milo{\v{s}} Ercegov{\v{c}}evi{\'c}, Martin Popel, and
  Omar~F Zaidan. 2011.
\newblock A grain of salt for the wmt manual evaluation.
\newblock In \emph{Proceedings of the Sixth Workshop on Statistical Machine
  Translation}, pages 1--11. Association for Computational Linguistics.

\bibitem[{Bojar et~al.(2016)Bojar, Graham, Kamran, and
  Stanojevi\'{c}}]{bojar-EtAl:2016:WMT2}
Ond\v{r}ej Bojar, Yvette Graham, Amir Kamran, and Milo\v{s} Stanojevi\'{c}.
  2016.
\newblock \href {http://www.aclweb.org/anthology/W16-2302} {Results of the
  wmt16 metrics shared task}.
\newblock In \emph{Proceedings of the First Conference on Machine Translation},
  pages 199--231, Berlin, Germany. Association for Computational Linguistics.

\bibitem[{Bryant et~al.(2017)Bryant, Felice, and
  Briscoe}]{bryant-felice-briscoe:2017:Long}
Christopher Bryant, Mariano Felice, and Ted Briscoe. 2017.
\newblock \href {http://aclweb.org/anthology/P17-1074} {Automatic annotation
  and evaluation of error types for grammatical error correction}.
\newblock In \emph{Proceedings of the 55th Annual Meeting of the Association
  for Computational Linguistics (Volume 1: Long Papers)}, pages 793--805,
  Vancouver, Canada. Association for Computational Linguistics.

\bibitem[{Bryant and Ng(2015)}]{bryant2015far}
Christopher Bryant and Hwee~Tou Ng. 2015.
\newblock How far are we from fully automatic high quality grammatical error
  correction?
\newblock In \emph{ACL (1)}, pages 697--707.

\bibitem[{Chen and Cherry(2014)}]{chen2014systematic}
Boxing Chen and Colin Cherry. 2014.
\newblock A systematic comparison of smoothing techniques for sentence-level
  bleu.
\newblock In \emph{Proceedings of the Ninth Workshop on Statistical Machine
  Translation}, pages 362--367.

\bibitem[{Choshen and Abend(2018{\natexlab{a}})}]{choshen2018conservatism}
Leshem Choshen and Omri Abend. 2018{\natexlab{a}}.
\newblock Inherent biases in reference-based evaluation for grammatical error
  correction and text simplification.
\newblock In \emph{Proceedings of the 56th Annual Meeting of the Association
  for Computational Linguistics (Volume 1: Long Papers)}.

\bibitem[{Choshen and Abend(2018{\natexlab{b}})}]{choshen2018usim}
Leshem Choshen and Omri Abend. 2018{\natexlab{b}}.
\newblock Reference-less measure of faithfulness for grammatical error
  correction.
\newblock In \emph{Proceedings of the 2018 Conference of the North American
  Chapter of the Association for Computational Linguistics: Human Language
  Technologies}.

\bibitem[{Dahlmeier and Ng(2012)}]{dahlmeier2012better}
Daniel Dahlmeier and Hwee~Tou Ng. 2012.
\newblock Better evaluation for grammatical error correction.
\newblock In \emph{Proceedings of the 2012 Conference of the North American
  Chapter of the Association for Computational Linguistics: Human Language
  Technologies}, pages 568--572. Association for Computational Linguistics.

\bibitem[{Dahlmeier et~al.(2013)Dahlmeier, Ng, and Wu}]{dahlmeier2013building}
Daniel Dahlmeier, Hwee~Tou Ng, and Siew~Mei Wu. 2013.
\newblock Building a large annotated corpus of learner english: The nus corpus
  of learner english.
\newblock In \emph{Proceedings of the Eighth Workshop on Innovative Use of NLP
  for Building Educational Applications}, pages 22--31.

\bibitem[{Dale et~al.(2012)Dale, Anisimoff, and Narroway}]{dale2012hoo}
Robert Dale, Ilya Anisimoff, and George Narroway. 2012.
\newblock Hoo 2012: A report on the preposition and determiner error correction
  shared task.
\newblock In \emph{Proceedings of the Seventh Workshop on Building Educational
  Applications Using NLP}, pages 54--62. Association for Computational
  Linguistics.

\bibitem[{Dras(2015)}]{dras2015evaluating}
Mark Dras. 2015.
\newblock Evaluating human pairwise preference judgments.
\newblock \emph{Computational Linguistics}, 41(2):337--345.

\bibitem[{Felice and Briscoe(2015)}]{felice2015towards}
Mariano Felice and Ted Briscoe. 2015.
\newblock Towards a standard evaluation method for grammatical error detection
  and correction.
\newblock In \emph{HLT-NAACL}, pages 578--587.

\bibitem[{Felice et~al.(2016)Felice, Bryant, and
  Briscoe}]{felice-bryant-briscoe:2016:COLING}
Mariano Felice, Christopher Bryant, and Ted Briscoe. 2016.
\newblock \href {http://aclweb.org/anthology/C16-1079} {Automatic extraction of
  learner errors in esl sentences using linguistically enhanced alignments}.
\newblock In \emph{Proceedings of COLING 2016, the 26th International
  Conference on Computational Linguistics: Technical Papers}, pages 825--835,
  Osaka, Japan. The COLING 2016 Organizing Committee.

\bibitem[{Graham et~al.(2012)Graham, Baldwin, Harwood, Moffat, and
  Zobel}]{graham2012measurement}
Yvette Graham, Timothy Baldwin, Aaron Harwood, Alistair Moffat, and Justin
  Zobel. 2012.
\newblock Measurement of progress in machine translation.
\newblock In \emph{Proceedings of the Australasian Language Technology
  Association Workshop 2012}, pages 70--78.

\bibitem[{Graham et~al.(2015)Graham, Baldwin, and Mathur}]{graham2015accurate}
Yvette Graham, Timothy Baldwin, and Nitika Mathur. 2015.
\newblock Accurate evaluation of segment-level machine translation metrics.
\newblock In \emph{Proceedings of the 2015 Conference of the North American
  Chapter of the Association for Computational Linguistics: Human Language
  Technologies}, pages 1183--1191.

\bibitem[{Grundkiewicz et~al.(2015)Grundkiewicz, Junczys-Dowmunt, Gillian
  et~al.}]{grundkiewicz2015human}
Roman Grundkiewicz, Marcin Junczys-Dowmunt, Edward Gillian, et~al. 2015.
\newblock Human evaluation of grammatical error correction systems.
\newblock In \emph{EMNLP}, pages 461--470.

\bibitem[{Koehn(2012)}]{koehn2012simulating}
Philipp Koehn. 2012.
\newblock Simulating human judgment in machine translation evaluation
  campaigns.
\newblock In \emph{International Workshop on Spoken Language Translation
  (IWSLT) 2012}.

\bibitem[{Krishna et~al.(2017)Krishna, Zhu, Groth, Johnson, Hata, Kravitz,
  Chen, Kalantidis, Li, Shamma et~al.}]{krishna2017visual}
Ranjay Krishna, Yuke Zhu, Oliver Groth, Justin Johnson, Kenji Hata, Joshua
  Kravitz, Stephanie Chen, Yannis Kalantidis, Li-Jia Li, David~A Shamma, et~al.
  2017.
\newblock Visual genome: Connecting language and vision using crowdsourced
  dense image annotations.
\newblock \emph{International Journal of Computer Vision}, 1(123):32--73.

\bibitem[{Kruskal and Sankoff(1983)}]{kruskal1983time}
Joseph~B Kruskal and David Sankoff. 1983.
\newblock \emph{Time Warps, String Edits, and Macromolecules: The Theory and
  Practice of Sequence Comparison}.
\newblock Addison-Wesley.

\bibitem[{Lopez(2012)}]{lopez2012putting}
Adam Lopez. 2012.
\newblock Putting human assessments of machine translation systems in order.
\newblock In \emph{Proceedings of the Seventh Workshop on Statistical Machine
  Translation}, pages 1--9. Association for Computational Linguistics.

\bibitem[{Mi{\l}kowski(2010)}]{milkowski2010developing}
Marcin Mi{\l}kowski. 2010.
\newblock Developing an open-source, rule-based proofreading tool.
\newblock \emph{Software: Practice and Experience}, 40(7):543--566.

\bibitem[{Napoles et~al.(2015)Napoles, Sakaguchi, Post, and
  Tetreault}]{napoles2015ground}
Courtney Napoles, Keisuke Sakaguchi, Matt Post, and Joel Tetreault. 2015.
\newblock Ground truth for grammatical error correction metrics.
\newblock In \emph{Proceedings of the 53rd Annual Meeting of the Association
  for Computational Linguistics and the 7th International Joint Conference on
  Natural Language Processing}, volume~2, pages 588--593.

\bibitem[{Napoles et~al.(2016{\natexlab{a}})Napoles, Sakaguchi, Post, and
  Tetreault}]{napoles2016gleu_update}
Courtney Napoles, Keisuke Sakaguchi, Matt Post, and Joel Tetreault.
  2016{\natexlab{a}}.
\newblock \href {http://arxiv.org/abs/1605.02592} {{GLEU} without tuning}.
\newblock \emph{eprint arXiv:1605.02592 [cs.CL]}.

\bibitem[{Napoles et~al.(2016{\natexlab{b}})Napoles, Sakaguchi, and
  Tetreault}]{napoles-sakaguchi-tetreault:2016:EMNLP2016}
Courtney Napoles, Keisuke Sakaguchi, and Joel Tetreault. 2016{\natexlab{b}}.
\newblock \href {https://aclweb.org/anthology/D16-1228} {There's no comparison:
  Reference-less evaluation metrics in grammatical error correction}.
\newblock In \emph{Proceedings of the 2016 Conference on Empirical Methods in
  Natural Language Processing}, pages 2109--2115. Association for Computational
  Linguistics.

\bibitem[{Ng et~al.(2014)Ng, Wu, Briscoe, Hadiwinoto, Susanto, and
  Bryant}]{ng2014conll}
Hwee~Tou Ng, Siew~Mei Wu, Ted Briscoe, Christian Hadiwinoto, Raymond~Hendy
  Susanto, and Christopher Bryant. 2014.
\newblock The conll-2014 shared task on grammatical error correction.
\newblock In \emph{CoNLL Shared Task}, pages 1--14.

\bibitem[{Papineni et~al.(2002)Papineni, Roukos, Ward, and
  Zhu}]{papineni2002bleu}
Kishore Papineni, Salim Roukos, Todd Ward, and Wei-Jing Zhu. 2002.
\newblock Bleu: a method for automatic evaluation of machine translation.
\newblock In \emph{Proceedings of the 40th annual meeting on association for
  computational linguistics}, pages 311--318. Association for Computational
  Linguistics.

\bibitem[{Sakaguchi et~al.(2016)Sakaguchi, Napoles, Post, and
  Tetreault}]{sakaguchi2016reassessing}
Keisuke Sakaguchi, Courtney Napoles, Matt Post, and Joel Tetreault. 2016.
\newblock Reassessing the goals of grammatical error correction: Fluency
  instead of grammaticality.
\newblock \emph{Transactions of the Association for Computational Linguistics},
  4:169--182.

\bibitem[{Sakaguchi et~al.(2014)Sakaguchi, Post, and
  Van~Durme}]{sakaguchi2014efficient}
Keisuke Sakaguchi, Matt Post, and Benjamin Van~Durme. 2014.
\newblock Efficient elicitation of annotations for human evaluation of machine
  translation.
\newblock In \emph{Proceedings of the Ninth Workshop on Statistical Machine
  Translation}, pages 1--11.

\bibitem[{Sennrich et~al.(2017)Sennrich, Firat, Cho, Birch, Haddow, Hitschler,
  Junczys-Dowmunt, L{\"a}ubli, Barone, Mokry et~al.}]{sennrich2017nematus}
Rico Sennrich, Orhan Firat, Kyunghyun Cho, Alexandra Birch, Barry Haddow,
  Julian Hitschler, Marcin Junczys-Dowmunt, Samuel L{\"a}ubli, Antonio
  Valerio~Miceli Barone, Jozef Mokry, et~al. 2017.
\newblock Nematus: a toolkit for neural machine translation.
\newblock \emph{arXiv preprint arXiv:1703.04357}.

\bibitem[{Sun and Zhou(2012)}]{sun2012joint_ibleu}
Hong Sun and Ming Zhou. 2012.
\newblock Joint learning of a dual smt system for paraphrase generation.
\newblock In \emph{Proceedings of the 50th Annual Meeting of the Association
  for Computational Linguistics: Short Papers-Volume 2}, pages 38--42.
  Association for Computational Linguistics.

\bibitem[{Xu et~al.(2016)Xu, Napoles, Pavlick, Chen, and
  Callison-Burch}]{Xu-EtAl:2016:TACL}
Wei Xu, Courtney Napoles, Ellie Pavlick, Quanze Chen, and Chris Callison-Burch.
  2016.
\newblock \href
  {https://cocoxu.github.io/publications/tacl2016-smt-simplification.pdf}
  {Optimizing statistical machine translation for text simplification}.
\newblock \emph{Transactions of the Association for Computational Linguistics},
  4:401--415.

\bibitem[{Yu et~al.(2017)Yu, Zhang, Wang, and Yu}]{yu2017seqgan}
Lantao Yu, Weinan Zhang, Jun Wang, and Yong Yu. 2017.
\newblock Seqgan: Sequence generative adversarial nets with policy gradient.
\newblock In \emph{AAAI}, pages 2852--2858.

\end{thebibliography}

\end{document}